\documentclass[]{fairmeta}

\newcommand{\methodname}{CoDi}
\usepackage{amsfonts}       
\usepackage{nicefrac}       
\usepackage{microtype}      
\usepackage{xcolor}         
\usepackage{tabularx}         
\usepackage{graphicx}
\usepackage{tikz}
\usepackage{pgfplots}
\usepackage{verbatim}

\title{CoDi: Conversational Distillation for Grounded Question Answering}

\author[1, *]{Patrick Huber}
\author[1]{Arash Einolghozati}
\author[1]{Rylan Conway}
\author[1]{Kanika Narang}
\author[1]{Matt Smith}
\author[1]{Waqar Nayyar}
\author[1]{Adithya Sagar}
\author[1]{Ahmed Aly}
\author[2, *]{Akshat Shrivastava}

\affiliation[1]{Reality Labs at Meta}
\affiliation[2]{FAIR at Meta}

\contribution[*]{Joint first author} 

\abstract{Distilling conversational skills into Small Language Models (SLMs) with approximately 1 billion parameters presents significant challenges. Firstly, SLMs have limited capacity in their model parameters to learn extensive knowledge compared to larger models. Secondly, high-quality conversational datasets are often scarce, small, and domain-specific. Addressing these challenges, we introduce a novel data distillation framework named \textbf{\methodname}~(short for  \textbf{Co}nversational \textbf{Di}stillation, pronounced ``Cody''), allowing us to synthesize large-scale, assistant-style datasets in a steerable and diverse manner.
Specifically, while our framework is task agnostic at its core, we explore and evaluate the potential of \methodname~on the task of conversational grounded reasoning for question answering. This is a typical on-device scenario for specialist SLMs, allowing for open-domain model responses, without requiring the model to ``memorize'' world knowledge in its limited capacity.
Our evaluations show that SLMs trained with \methodname-synthesized data achieve performance comparable to models trained on human-annotated data in standard metrics. Additionally, when using our framework to generate larger datasets from web data, our models surpass larger, instruction-tuned models in zero-shot conversational grounded reasoning tasks.}

\date{\today}
\correspondence{Patrick Huber at \email{patrickhuber@meta.com}}

\begin{document}

\maketitle

\section{Introduction}
\label{section:intro}


Small Language Models (SLMs), defined here as having approximately 1 billion parameters, do not exhibit the same level of generalization and emergent abilities as Large Language Models (LLMs) \citep{fu2023specializing}. For instance, when the model size is reduced from 70 billion to 7 billion and further to 1.3 billion parameters, performance on the "Massive Multitask Language Understanding" (MMLU) benchmark decreases by 23.6\% and 43.2\% respectively \citep{touvron2023llama, xia2023sheared}.  These sharply declining results of general language understanding across model sizes highlight the limitations of SLMs as generalist language models. As a result, we believe that smaller language models are more suited as task specialists, solving a small subset of tasks.


Within this space of task specialist models, conversational abilities of SLMs play a crucial role for their deployment on edge devices, such as mobile phones and wearables. Despite the benefits of on-device models (e.g., latency and availability), enabling SLMs to be conversational agents is a difficult problem due to (1) the limited capacity to learn and retain extensive knowledge in the parameters of SLMs and (2) the lack of large-scale, high quality, conversational datasets, which are often limited to narrow domains \citep{duan2023botchat}. 

The latter point is thereby largely attributed to the notoriously difficult human annotation of multi-turn data. Manually creating conversational multi-turn datasets is resource intensive, since every sample requires a valid and meaningful conversational history. 
Hence, annotating multi-turn datasets is either significantly more resource-intensive (at the same scale) or results in much smaller datasets using the same resources, e.g. CoQA \citep{reddy2019coqa}, QuAC \citep{choi2018quac}, or OpenAssistant \citep{köpf2023openassistant}. This, in turn, leads to a shortage of suitable training resources for conversational models in terms of volume and diversity. 

In this paper, we argue that the limited amount of human-annotated (i.e. ``gold'') conversational data should be used primarily for evaluation purposes, creating a void in the training space for conversational multi-turn models. Aiming to fill this gap, we propose \textbf{Co}nversational \textbf{Di}stillation (\textbf{\methodname}) as a path forward to synthesize diverse and steerable multi-turn conversations from black-box ``teacher'' LLMs. Using this new source of high-quality, high-quantity conversational data, we knowledge distill conversational abilities into on-device sized ``student'' SLMs.


While the \methodname~framework itself is task agnostic and can be applied to a large variety of diverse tasks, we show its application to the challenging task of grounded reasoning in this paper. Evaluating the framework in this natural multi-turn setting, we find that our distillation approach shows competitive results to models fully fine-tuned on human-annotated datasets and consistently outperforms similar-sized instruction-tuned baselines in zero-shot settings.

\begin{figure*}[h]
    \centering
    \includegraphics[width=\linewidth]{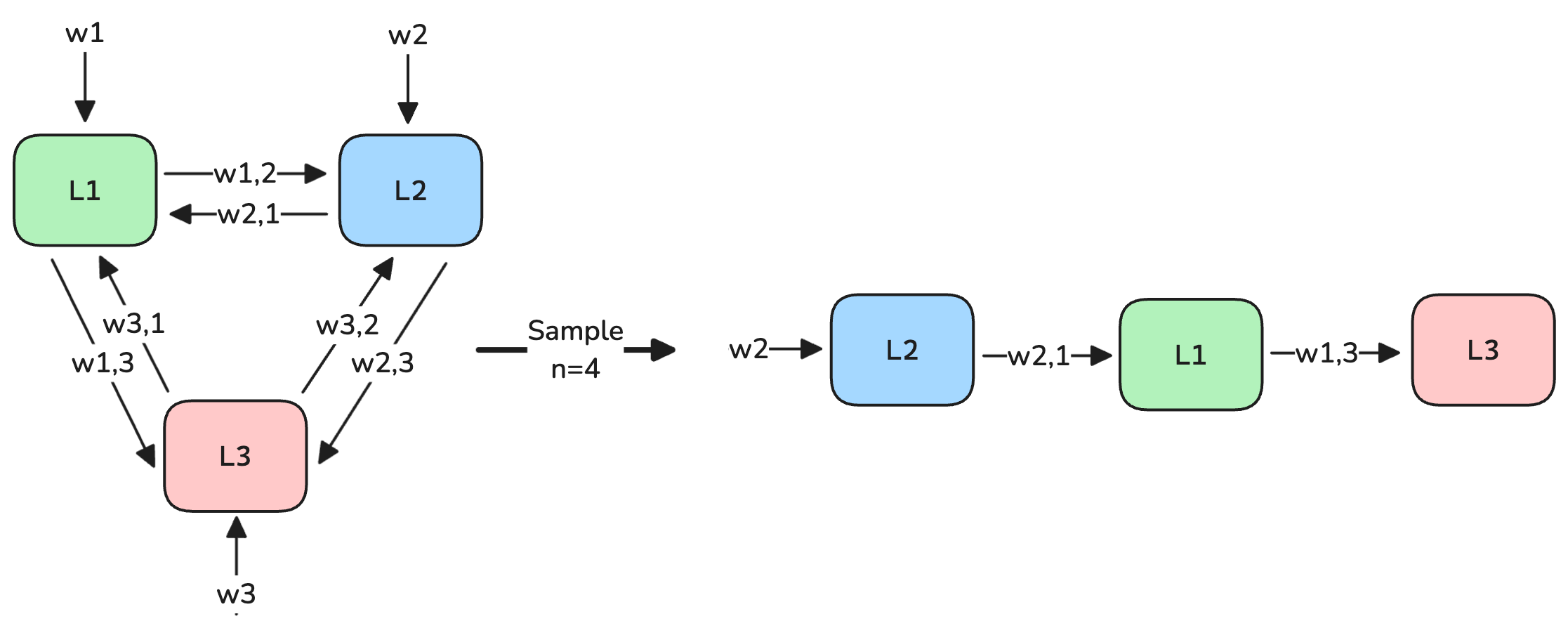}
    \caption{Conversational Graph Generation Example. Left: General conversational graph, Right: Rolled out version of a sampled graph at length n=4}
    \label{fig:conv_graph}
\end{figure*}

\section{Method}
\label{sec:methodc}
In order to enrich Small Language Models (SLMs), a popular approach has been to ``black-box'' distill capabilities from large language models \citep{elmadany-etal-2023-orca, hsieh2023distilling, gunasekar2023textbooks} through data augmentation or synthesis. Following this intuition, we propose a new distillation methodology
to enrich small language models with conversational grounded reasoning abilities. In comparison to existing distillation paradigms mostly focusing on synthesized data in the instruction-tuning domain (e.g. WizardLM \citep{xu2023wizardlm}, Alpagasus \citep{chenalpagasus}, and Alpaca \citep{alpaca}), we focus on synthesizing true multi-turn conversations using a ``turn-by-turn'' generation paradigm to target the shortcoming of current LLMs when used at the conversation level. This way, we aim to increase the distillation performace 
along two dimensions: diversity and steerability.
%

\begin{figure}[t]
    \centering
    \includegraphics[width=0.65\linewidth]{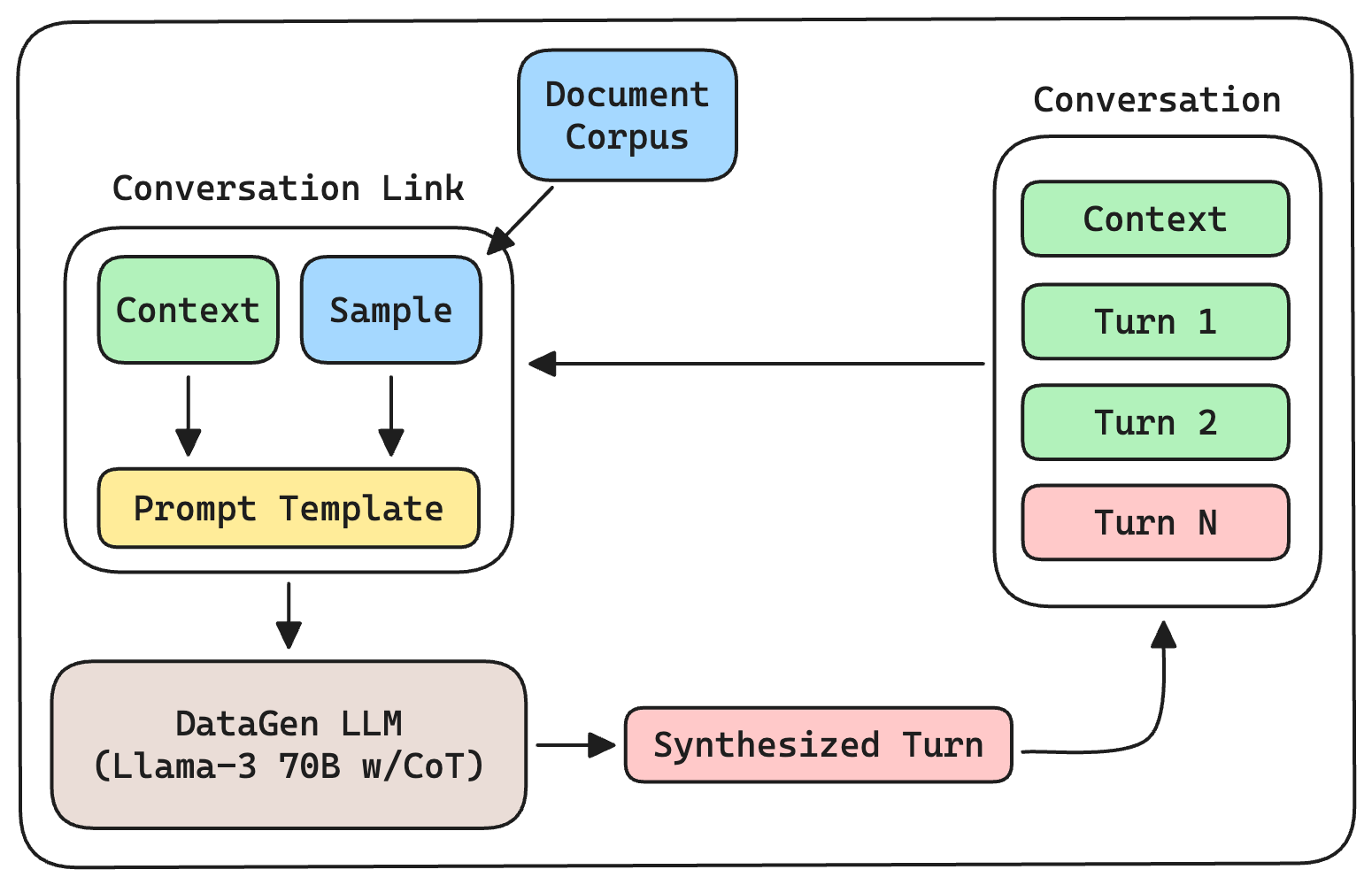}
    \caption{A single generation step to synthesize a new turn in the conversation.}
    \label{fig:generation_UML}
\end{figure}

\subsection{Distilling Conversational Abilities}
\label{sec:method:distillation}

Enabling conversational abilities in SLMs through supervised fine-tuning is a data intensive process requiring careful data curation at scale. 
Multiple lines of research, e.g. \citet{hsieh2023distilling, mukherjee2023orca, gunasekar2023textbooks, zhou2023lima} recently found that data quality, besides scale, is imperative when aiming to effectively distill language models.
Building on top of these insights, we present a distillation method to explicitly enrich SLMs with conversational reasoning abilities by introducing three new concepts to generate diverse, engaging and fluent conversations: conversational graphs, turn based prompt augmentations, and explicit linguistic features. 


\subsubsection{Conversational Graph Generation}
\label{sec:conv_graph}
Synthesizing diverse, yet naturally flowing conversations is imperative when imitating human interactions. Generating valid, diverse and coherent conversation is hence a key challenge for our multi-turn synthesis framework. To to able to make some guarantees regarding conversation validity, we propose a conversational graph generation approach inspired by Markov Chains. An example conversation graph is shown in Figure~\ref{fig:conv_graph}. While specific instances of the conversational graph vary based on the synthesis task at hand, the general structure defines a set of ``conversation links'' (visualized as vertices), representing blueprints for prompting conversational turns, connected by transition probabilities (here: edges) used to sample a ``conversational chain'', a sequence of conversation links representing a valid and natural multi-turn conversation from the graph. 

In general, for a specific task (e.g. grounded question answering), a conversational graph $G = (V, E)$ with vertices $V = (\emptyset, L_1, L_2, L_3)$ (defining ``conversation links'') and edges $E = (w_{\emptyset, 1}, w_{1, 2}, w_{2, 1}, ...)$ (representing transition probabilities) is defined. To generate a question-answering conversation using $G$, we sample from the set of valid edges (e.g. \{$w_{\emptyset, 1}$, $w_{\emptyset, 2}$ and $w_{\emptyset, 3}$\} if $V=\emptyset$) and instantiate the sampled target link (e.g. $L_2$ in Figure~\ref{fig:conv_graph}). In the next step, we now sample from \{$w_{2, 1}$, $w_{2, 3}$\}, since $V=L_2$). 
Once we reach the defined conversation length $n$ (e.g. 4 in Figure~\ref{fig:conv_graph}), we end the graph traversal and return the valid and diverse conversational blueprint for teacher model synthesis.



\begin{figure}[h]
    \centering
    \includegraphics[width=0.7\linewidth]{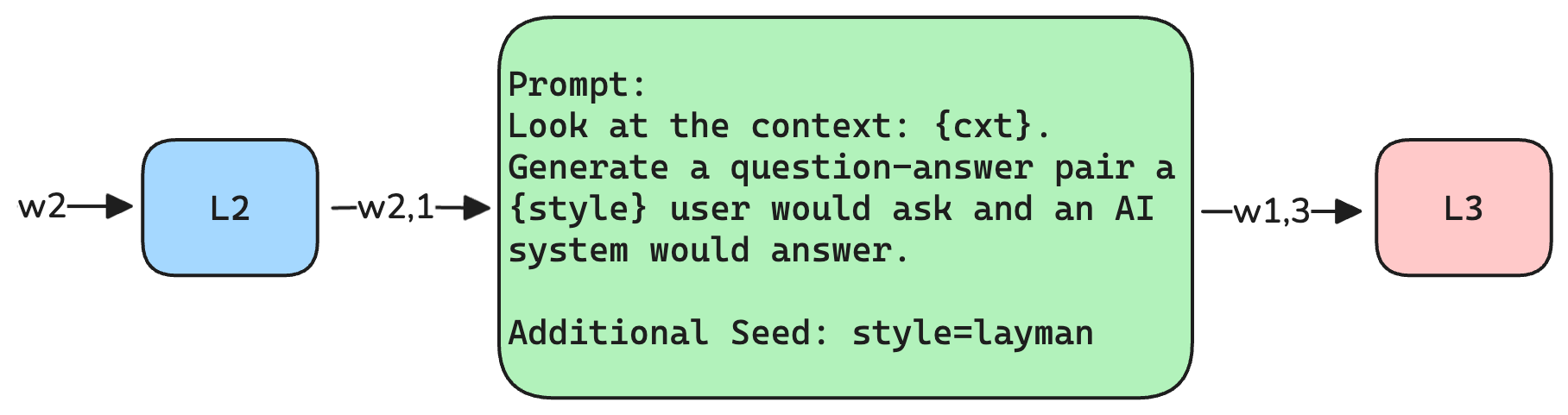}
    \caption{Per-turn conversational link augmentation with prompt and (optional) seed data}
    \label{fig:turn_based_prompts}
\end{figure}

\subsubsection{Conversational Links}
\label{sec:intra_turn}
Once a valid conversational chain is sampled from the graph according to the transition probabilities, the links are executed in-order (e.g., in Figure~\ref{fig:conv_graph}: $L_2\rightarrow L_1 \rightarrow L_2 \rightarrow L_3$), with an optional context prepended to the sequence for grounded tasks (e.g. grounded reasoning). While the conversational graph defines the ``macro'' level conversation structure, the individual links define the conversational turns themselves. Each link 
thereby contains: (1) A link-specific prompt to steer the conversation and (2) potential seed data for the current step (see Figure \ref{fig:turn_based_prompts}).
The prompt is mandatory in every link and should ideally utilize methods to enhance the distillation quality, such as ``Chain-of-Thought'' reasoning steps (CoT, \citet{wei2023chainofthought}, not shown in Figure~\ref{fig:turn_based_prompts}). The prompt can optionally (depending on the link itself) utilize additional seed data samples to support diversity in the conversational chain or, in cases where auxiliary data is required, directly use external information (e.g., context). The role of the prompt and seed are configurable depending on the link and present the primary means to enable explicit conversational phenomena in the conversation.

Figure \ref{fig:generation_UML} shows the flow of data in a single generation step, where the prompt template, originating in the Link, is filled in by the conversational context and diversified by an optional data point. It is then used to generate a new conversational turn through the teacher ``DataGen LLM''. The newly generated conversation turn is then added to the stored conversation and the process is started over with the next link in the conversational chain.



\begin{figure}[h]
    \centering
    \includegraphics[width=\linewidth]{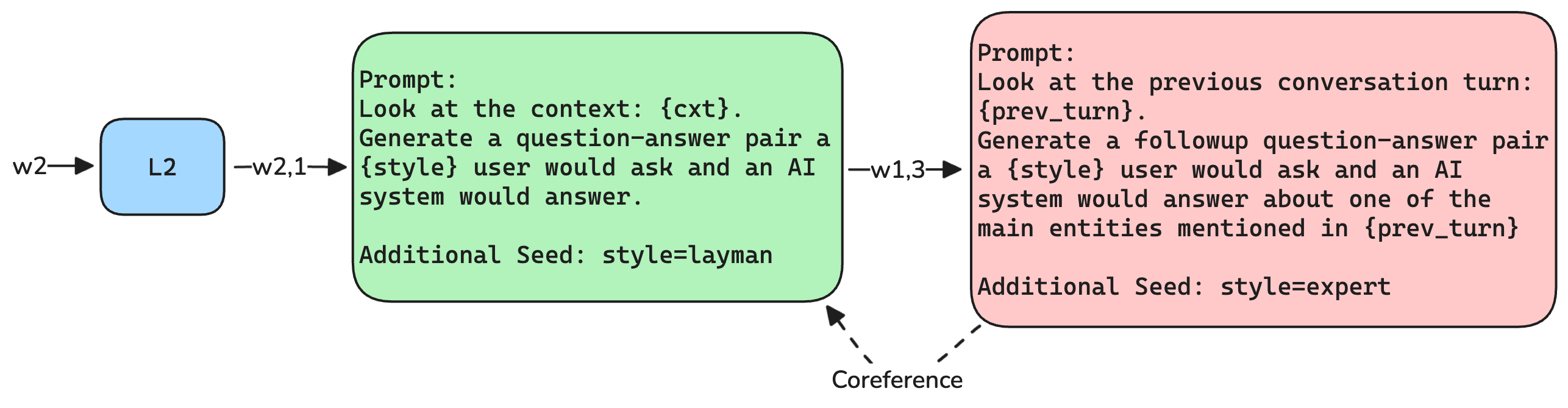}
    \caption{Example of linguistic phenomena used in the final turn prompt.}
    \label{fig:conv_phenom}
\end{figure}

\subsubsection{Linguistic Phenomena}
\begin{figure}
    \centering
    \includegraphics[width=\linewidth]{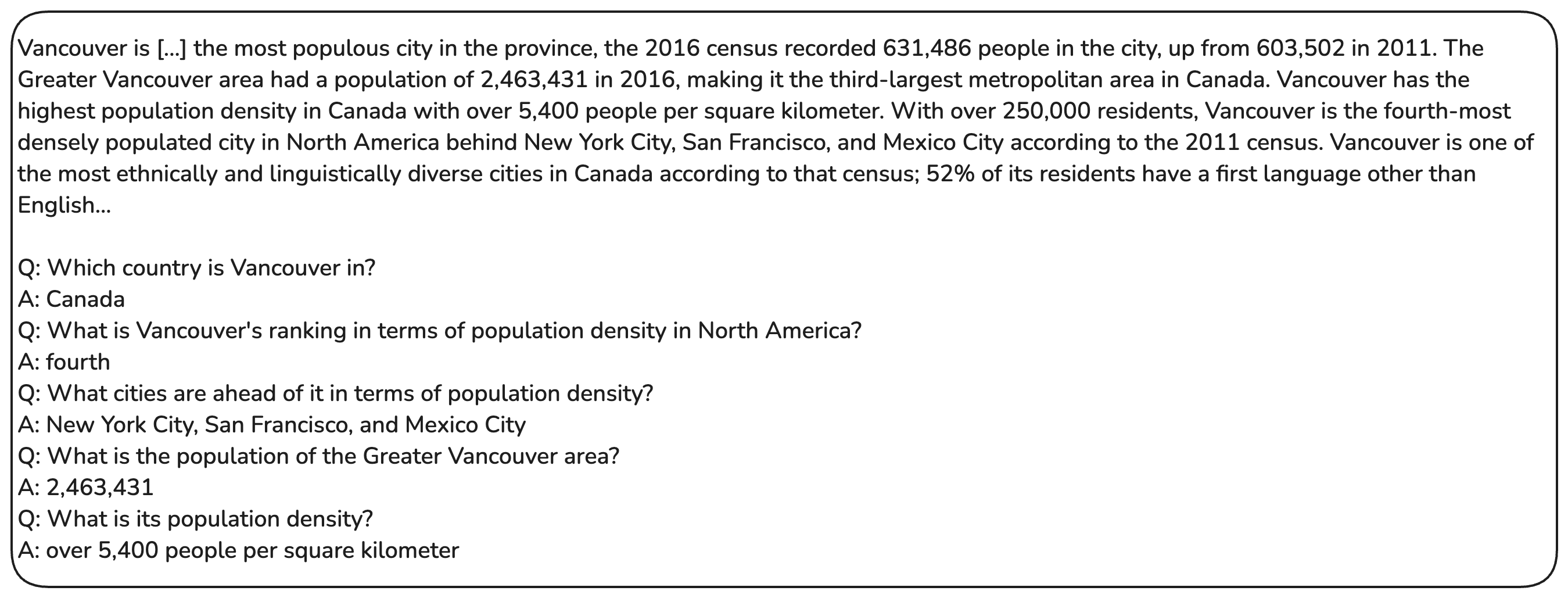}
    \caption{\methodname~synthesized example. Context document taken from the CoQA training corpus.}
    \label{fig:ex_1}
\end{figure}

Up to this point, our generations could still result in a sequence of independent, single turn utterances. Inspired by everyday conversations between humans, we use explicit linguistic phenomena to naturally tie turns together. Figure~\ref{fig:conv_phenom} visualizes this core principle behind the approach for our running example using coreference to refer back to a prior conversational turn. This way, we explicitly tie together entity mentions in the context and synthesize semantic follow-ups. Figure~\ref{fig:ex_1} shows an example of a synthesized multi-turn conversation based on a CoQA context document.


This results in our final, steerable and diverse conversation synthesis framework proposed in this paper.

\subsection{Small Scale Conversational Specialists} 
\label{sec:slm_training}
To enable Small Language Models (SLMs) to acquire conversational abilities, we introduce a new data format, which adds additional flexibility to the otherwise limited set of roles in common chat templates, such as the Llama format \citep{touvron2023llama}. Specifically, we allow for an arbitrary number of conversational roles. In the most basic case, a conversation contains two roles, a \textit{USER} and an \textit{AGENT} (where even this naming convention is flexible) as shown below:

\begin{verbatim}
[USER] What color is the sky? [/USER]
[AGENT] The sky is blue. [/AGENT]
\end{verbatim}

Besides the core roles of \textit{USER} and \textit{AGENT}, the \textit{CONTEXT} role is a common third role, sometimes also referred to as the \textit{SYSTEM} role. The \textit{CONTEXT} represents any contextual information the model requires to reason over, which can range anywhere from an instruction to external knowledge in the form of documents, (external) conversations, structured representations, or additional modalities (e.g., images). The \textit{CONTEXT} generally plays a crucial role as the interface between on-device information and the model, as shown below:

\begin{verbatim}
[CONTEXT] <USER MESSAGE INBOX> [/CONTEXT]
[USER] Do I have any messages? [/USER]
[AGENT] You have 2 new messages from Alex. He asks about your weekend plans. Reply? [/AGENT]
\end{verbatim}

The second advantage to our approach is Role Weighting,
which refers to the ability to adjust the training loss depending on the role. For example, the primary responsibility of an assistant-style model is to generate meaningful \textit{AGENT} turns based on a context. To emphasize this behavior, learning the \textit{AGENT} role should be the primary objective during model training. 
In general, having the ability to control the loss based on the underlying role allows us to gain better control over the learning process, leading to more well defined models.


\section{Experiments}
\label{sec:experiments}

 
\subsection{Task}
\label{sec:task}
To evaluate our distillation approach for small-scale conversational agents we choose the task of grounded reasoning for a variety of reasons. First, targeting smaller, specialist language models at on-device size, a natural focus is put on the models interaction with on-device context (e.g. summarize notes on-device). Thus, a knowledge grounded task is a natural choice. Second, given on-device limitations (which restrict usage to SLMs), 
we can not expect the model to reliably retain large amounts of information. This is evident when evaluating SLMs on knowledge intensive tasks, e.g., MMLU \citep{touvron2023llama, xia2023sheared}. To this end, a grounded agent offers a more attainable goal. We therefore focus on question answering centered scenarios, where users interact with the system and expect truthful (i.e. grounded) responses to their requests. 


\subsection{Models}
\label{sec:models}
In this paper, we use two main models:\\
\textbf{The teacher model (LLM)} is the 70B Llama3 instruction tuned checkpoint, one of the most capable open-source, instruction tuned large language models to date. \\ 
\textbf{The student model (SLM)} is a pre-trained 1.4B Llama2-style model.



We use additional model sizes and architectures for ablation experiments and comparisons. Specifically, we further employ the 70B Llama2 instruction tuned checkpoint as an alternative teacher model \citep{touvron2023llama}. As an alternative student model, we use a pre-trained 500M Llama2-style model. As instruction-tuned baselines, we explore 7B \citep{touvron2023llama}, 1.4B and 500M Llama2-style models as well as Phi-3 \citep{abdin2024phi3}. Further baselines are taken from the literature and directly cited in the relevant sections.

\begin{figure}[h]
    \centering
    \includegraphics[width=.8\linewidth]{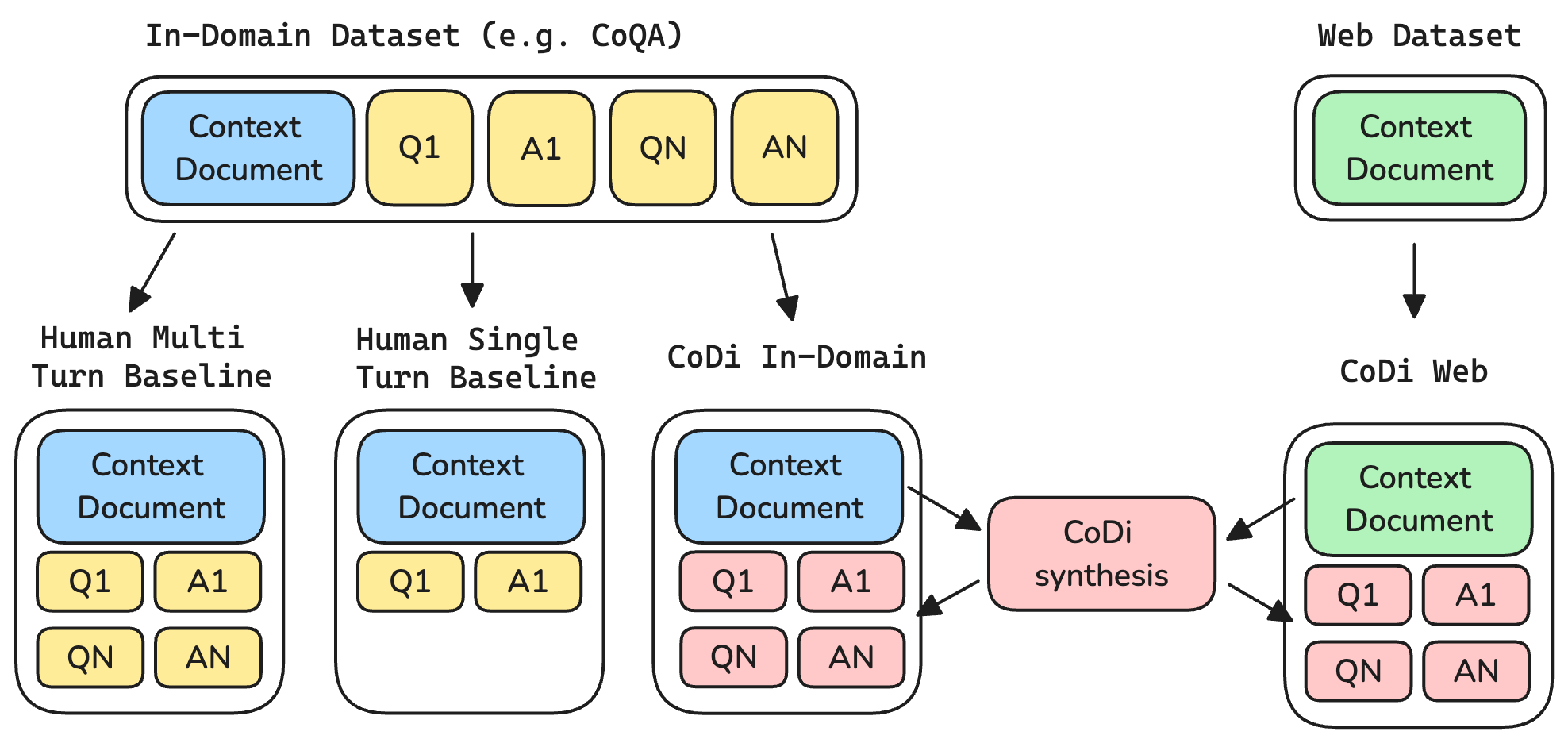}
    \caption{Training setup of human annotated baselines and in-domain / web-based \methodname~versions.}
    \label{fig:evaluation_process}
\end{figure}

\subsection{Datasets}
\label{sec:datasets}
We utilize two sets of datasets in this work: First, to generate diverse data for our \methodname~distillation approach, we present our synthesis datasets in section~\ref{sec_synth_data}. Subsequently, we describe our evaluation datasets to compare our distilled models against supervised and zero-shot baselines in section~\ref{sec_eval_data}. Lastly, section \ref{sec_metrics} describes the dataset metrics used for our comparisons.

\subsubsection{Synthesis Datasets}
\label{sec_synth_data}
To evaluate the grounded reasoning abilities of our \methodname~distillation approach, we explore two synthesis scenarios: intra-domain and zero-shot.

\begin{table}[h]
    \centering
    \begin{tabular}{l|r|r}
       Dataset & CoQA & QuAC \\
       \hline
       Dialogs & 8,199 & 13,594 \\
       Questions & 121,300 & 98,407 \\
       Passage length & 270 & 401 \\
       Avg. Turns & 15 & 7 \\
    \end{tabular}
    \caption{Dataset dimensions of the intra-domain grounded reasoning training portion.}
    \label{tab:gqa_dataset_dimensions}
\end{table}

\paragraph{Intra-domain Synthesis:}
We aim to compare our distilled models (based on purely synthetic conversations) against gold datasets, using the same set of available documents during training (see in-domain contexts (blue) in Figure \ref{fig:evaluation_process}). Specifically, we use two common multi-turn grounded question answering tasks:
Conversational Question Answering (CoQA) \citep{reddy2019coqa} and Question Answering in Context (QuAC) \citep{choi2018quac}. Both tasks test the ability of a system to generate responses to a query based on a given context and conversational history. The dataset statistics are shown in Table~\ref{tab:gqa_dataset_dimensions}. While both datasets target conversational question-answering, CoQA answers are generally in a factoid-style (i.e. short and precise), while QuAC responses are more elaborate and, as a result, longer. We call our intra-domain distilled model \textbf{``\methodname~In-Domain''}.

\paragraph{Zero-shot Synthesis:}
As shown in the bottom right corner in Figure \ref{fig:evaluation_process}, in this setting, we use web documents as the context to reason over in question answering style. 
With the automatic nature of our approach to constructing conversations, we are only limited by the amount of available seed data to ensure data diversity. As a result, we can leverage large-scale data collections to effectively scale our \methodname~approach across several orders of magnitude. To show the effect of scaling up our data distillation approach, we use readily available web data, hence calling this version of our models \textbf{``\methodname~Web''}



\subsubsection{Evaluation Datasets}
\label{sec_eval_data}
We evaluate all models on two human test sets provided as part of the CoQA \citep{reddy2019coqa} and QuAC \citep{choi2018quac} corpora\footnote{As is common practice, we use the validation portion of the datasets as our evaluation set, while sampling a validation set from the original training split.}. We utilize the original context as well as the complete grounded question-answering conversation, akin to the human multi-turn training set shown in the top left in Figure \ref{fig:evaluation_process}.
As a held out, zero-shot evaluation scenario, we show results on the (grounded) abstract summarization task based on two popular summarization datasets: CNN/DM \citep{nallapati2016abstractive} and XSum \citep{xsum-emnlp}.

\subsubsection{Evaluation Metrics} 
\label{sec_metrics}
Due to the nature of LLM synthesized conversations, our generations are more aligned with the long-form responses in the QuAC dataset, than the factoid style CoQA answers. For this reason, we decide to 
opt for the less strict recall metric for the CoQA dataset to not overly penalize results based on their output length. 
To ensure that model candidates don't exploit the recall metric by generating excessively long responses, we supplement our results with additional response length statistics. Given the similar human-like style of QuAC gold answers and our generations, we present the common F1-scores here.

Given our focus on conversational grounded abilities, we further evaluate all models in two distinct scenarios: (1) The standard approach, assuming a gold conversation history for every turn in the conversation and 
(2) a more realistic scenario, in which we evaluate the ability of the model to predict the conversation as a whole by using prior SLM predictions as the conversation history.


\subsection{Baselines}
\label{sec:baselines}

\paragraph{Human Single Turn:} This baseline uses the in-domain, single-turn, human-annotated question answering dataset. To obtain this baseline, we use the original multi-turn human-annotated dataset and remove all turns past the first interaction (see top right in Figure \ref{fig:evaluation_process}). 
We expect multi-turn datasets to exceed the performance of this baseline on conversational tasks. 

\paragraph{Human Multi Turn:} Similar to Single-Turn Human annotations, this baseline consists of the original, human multi-turn conversations provided as part of the CoQa and QuAC datasets (see top left in Figure \ref{fig:evaluation_process}). This baseline is a super set of the setup described above and predicted to  
be significantly better for tasks that required conversational reasoning.


\paragraph{Instruction-Tuned:} A common alternative to task specialist models are instruction-tuned checkpoints. Given that these models are solely prompted to solve a specific task (e.g., as done in \citet{liu2024chatqa}), instruction-tuned models are an important alternative to task specialists in domains where training data is sparse or non-existent. We use a range of instruction-tuned baselines at different parameter sizes to verify the benefit of our \methodname~framework.



\begin{table}[h]
    \centering
    \begin{tabular}{l|r|rr|rr}
        Eval Dataset && \multicolumn{2}{c|}{CoQA} & \multicolumn{2}{c}{QuAC} \\
        \hline
        Metric & \#Params & \multicolumn{2}{c|}{Recall} & \multicolumn{2}{c}{F1-score} \\
        \hline
        Context && Gold & Pred & Gold & Pred \\
        \hline
        \multicolumn{6}{c}{In-Domain} \\
        \hline
        Human Single Turn & 1.4B & 77.3 & 73.7 & 36.73 & 30.80 \\
        \methodname~In-Domain & 1.4B & 84.5 & 81.8 & 40.97 & 38.42 \\
        Human Multi-Turn & 1.4B & \textbf{85.2} & \textbf{82.5} & \textbf{47.20} & \textbf{41.02} \\
        \hline
        \multicolumn{6}{c}{Zero-Shot} \\
        \hline
        Instruction-Tuned & 1.4B & 79.7 & 68.9 & 21.37 & 18.04 \\
        \methodname~Web & 1.4B & \textbf{86.3} & \textbf{84.2} & \textbf{38.66} & \textbf{35.51} \\
        \hline
        Phi-3 & 3.8B & 89.0 & 78.8 & 34.56 & 16.24 \\ 
        \hline
        Instruction-Tuned & 7B & 85.0 & 82.8 & 25.99 & 17.58 \\
        \methodname~Web & 7B & \textbf{91.0} & \textbf{89.3} & \textbf{39.63} & \textbf{37.41} \\
        \hline
        Instruction-Tuned$\dagger$ & 70B & -- & -- & 32.47 & -- \\
    \end{tabular}
    \caption{GQA In-Domain and Zero-Shot Results on CoQA and QuAC. Llama2 based, if no model mentioned, $\dagger$ = Taken from \citet{liu2024chatqa}}
    \label{tab:gqa_results_zs}
\end{table}

\section {Results}
\label{sec:results}





The main results of this paper are presented in Table \ref{tab:gqa_results_zs}, containing two sections: 

At the top, we show the intra-domain setting, where we synthesize grounded conversations based on documents from the CoQA and QuAC training sets and evaluate the conversational question answering ability on the respective test sets (also see blue ``in domain'' examples shown in Figure \ref{fig:evaluation_process}). The goal of this experiment is to see if our automatically synthesized data can close the gap between human-annotated single-turn and multi-turn baselines using LLM distillation. Specifically, if the distilled model does not significantly outperform the single-turn human baseline, the generated conversations are likely not useful for conversational question answering tasks. Similarly, if using our synthesized conversational dataset as the training corpus results in similar performance compared to the human multi-turn training dataset, we can conclude that the conversational data synthesis and distillation approach is indeed effective. Looking at the top portion of Table \ref{tab:gqa_results_zs} 
we see that we are able to close the single-turn -- multi-turn gap by 92\% (91\% with gold context) and 75\% (40\% with gold context) for CoQA and QuAC respectively, despite using exclusively synthesized conversations. 

In the second set of results, (bottom of Table \ref{tab:gqa_results_zs}) we show zero-shot comparisons of our large-scale synthesized and distilled \methodname~Web model (trained on one million generated conversations) against multiple baselines. Specifically, we compare the performance against instruction-tuned Llama2-like baselines of various sizes \citep{touvron2023llama}, as well as the 3.8B parameter Phi-3 model \citep{abdin2024phi3}. As shown in Table \ref{tab:gqa_results_zs}, we find that \methodname~Web consistently outperforms instruction-tuned baselines at similar scale and above. For example, the 1.4B \methodname~Web model outperforms the 1.4B and 7B instruction-tuned model, while \methodname~Web 7B improves performance compared to the 7B and 70B instruction-tuned models. We further outperform the Phi-3 baseline (2x larger) in three of the four settings.

Comparing the zero-shot synthesis performance against the human multi-turn model (shown in the top section of Table \ref{tab:gqa_results_zs}), we find that our \methodname~Web model improves over the baseline on the CoQA dataset, while under performing on QuAC. 
Taking a closer look at the performance gap between the conversation history settings (i.e. ``Gold'' and ``Pred''), we observe a consistent degradation for the predicted conversation history. Despite being present in all evaluations, the gap is generally smaller in our \methodname~models compared to human baselines and other zero-shot approaches. This points towards a more coherent conversational trajectory across multi-turn conversations when using our distillation approach.

Given these results, we make a clear case for our conversational \methodname~framework as an effective method to improve task-specific model performance for grounded reasoning tasks. Lastly, to ensure the recall results we show in Table \ref{tab:gqa_results_zs}) are valid, 
we present a supplementary word count analysis for the model responses in Table \ref{tab:gqa_response_length}, confirming that our distillation approach does not exploit the recall metric through excessively long generations.



\begin{table}[h]
    \centering
    \begin{tabular}{l|r|rr|rr|rr}
        Model & \# Params & \multicolumn{2}{c|}{Average} & \multicolumn{2}{c|}{Median} & \multicolumn{2}{c}{90P} \\
        \hline
        Context & & Gold & Pred & Gold & Pred & Gold & Pred \\
        \hline
        \multicolumn{7}{c}{In Domain} \\
        \hline
        Human Single Turn & 1.4B & 2.1 & 2.2  &  1 & 1  &  4 & 4 \\
        Human Multi Turn & 1.4B & 2.4 & 2.5  &  2 & 2  &  5 & 5 \\
        \methodname~In-Domain & 1.4B & 4.8 & 4.9  &  3 & 3  &  10 & 11 \\
        \hline
        \multicolumn{7}{c}{Zero Shot} \\
        \hline
        \methodname~Web & 1.4B & 4.7 & 4.7  &  3 & 3  &  11 & 10 \\
        Instruction-Tuned & 1.4B & 65.9 & 29.5  &  28 & 7  &  154 & 117 \\
        Phi-3 & 3.8B & 89.1 & 16.3  &  49 & 5  &  209 & 27 \\
        Instruction-Tuned & 7B & 12.2 & 7.6  &  7 & 3  &  25 & 19 \\
        \hline
        Gold Answer & -- & 2.52 &--& 2 &--& 5 & -- \\
    \end{tabular}
    \caption{GQA Response length in words for CoQA. Llama2 based, if no model mentioned}
    \label{tab:gqa_response_length}
\end{table}

\section {Analysis}
\label{sec:analysis}
To better understand the abilities and potential shortcomings of our novel distillation approach, we're now exploring a range of additional ablation experiments. We explore seven main dimensions of the approach to understand the role of dataset scale, different student model sizes, teacher models and the per-turn performance. We then ensure reasonable language modeling evaluations, show a small scale human evaluation as well as zero-shot performances on tasks besides grounded question answering: abstractive summarization.  

For all ablations shown below, we use 10,000 conversation samples synthesized using Llama3 70B Instruct from our zero-shot web seed and the Llama2-like 1.4B student, unless stated otherwise.

\subsection{Distillation Scale Comparison}
This experiment targets one of the most important properties of our \methodname~distillation framework: the ability to synthesize data at scale. While small-scale, human-annotated datasets exist, producing them is resource-intensive. Using \methodname, we can synthesize large amounts of diverse conversational data across multiple orders of magnitude. In the ablation experiment in Table \ref{tab:scale} we show the influence of the number of synthesized conversation on the model performance. The trend across orders of magnitude is clear: larger synthesis scales improve the performance near linearly up to 1 million samples.

\begin{table}[h]
    \centering
    \begin{tabular}{l|r|rr|rr}
        Eval Dataset & & \multicolumn{2}{c|}{CoQA} & \multicolumn{2}{c}{QuAC} \\
        \hline
        Metric & \# Params & \multicolumn{2}{c|}{Recall} & \multicolumn{2}{c}{F1-score} \\
        \hline
        Context & & Gold & Pred & Gold & Pred \\
        \hline
        \methodname~Web 10k & 1.4B & 80.8 & 78.4 & 36.69 & 31.50 \\
        \methodname~Web 100k & 1.4B & 83.1 & 80.9 & 37.51 & 33.92 \\
        \methodname~Web 1M & 1.4B & \textbf{86.3} & \textbf{84.2} & \textbf{38.66} & \textbf{35.51} \\
    \end{tabular}
    \caption{GQA Zero-Shot Results on CoQA and QuAC across four different synthesis scales. Llama2 based, if no model mentioned.}
    \label{tab:scale}
\end{table}

\subsection{Student Model Size Comparison}
This ablation explores the impact of different student model sizes on the conversational question answering performance. We compare the Llama2-like 1.4B student checkpoint used in the main results (using the full dataset) with a Llama2-like 500M student checkpoint. In Table \ref{tab:model_size_ablation} we observe (as expected) a clear quality regression when moving from 1.4B parameters to the 500M size. However, compared to the human-annotated training datasets (see ``Human multi turn''), our distilled zero-shot model still only slightly under performs the model trained on human multi-turn data.

\begin{table}[h]
    \centering
    \begin{tabular}{l|r|rr|rr}
        Eval Dataset & & \multicolumn{2}{c|}{CoQA} & \multicolumn{2}{c}{QuAC} \\
        \hline
        Metric & \# Params & \multicolumn{2}{c|}{Recall} & \multicolumn{2}{c}{F1-score} \\
        \hline
        Context & & Gold & Pred & Gold & Pred \\
        \hline
        Human Single Turn & 500M & 55.5 & 52.2 & 29.28 & 25.50 \\
        \methodname~Web & 500M & 72.3 & 70.1 & 30.76 & 29.06 \\
        Human Multi Turn & 500M & \textbf{72.9} & \textbf{70.9} & \textbf{39.12} & \textbf{32.58} \\
        \hline
        \methodname~Web & 1.4B & 80.8 & 78.4 & 36.69 & 31.50 \\
    \end{tabular}
    \caption{GQA Zero-Shot Results on CoQA and QuAC using the 500M Llama2-like model on the full (1M) dataset.}
    \label{tab:model_size_ablation}
\end{table}

\subsection{Teacher Model Comparison}
This experiment explores the impact of the teacher model synthesis quality on the final distillation performance. We compare the 70B instruction-tuned Llama2 and Llama3 models. In Table \ref{tab:teacher_model_ablation} we see a clear quality improvement from using the Llama3 teacher model, showing a 2\%+ absolute performance improvement of the distilled student models when trained on Llama3 synthesized conversations.

\begin{table}[h]
    \centering
    \begin{tabular}{l|r|rr|rr}
        Eval Dataset & & \multicolumn{2}{c|}{CoQA} & \multicolumn{2}{c}{QuAC} \\
        \hline
        Metric & \# Params & \multicolumn{2}{c|}{Recall} & \multicolumn{2}{c}{F1-score} \\
        \hline
        Context & & Gold & Pred & Gold & Pred \\
        \hline
        Llama2 Instruction-Tuned & 70B & 77.40 & 76.50 & 31.86 & 29.51 \\ 
        Llama3 Instruction-Tuned & 70B & \textbf{80.80} & \textbf{78.40} & \textbf{36.69} & \textbf{31.50} \\ 
    \end{tabular}
    \caption{GQA Zero-Shot Results on CoQA and QuAC using Llama2 and Llama3 synthesized distillation data.}
    \label{tab:teacher_model_ablation}
\end{table}


\subsection{Per-Turn Performance Comparison}
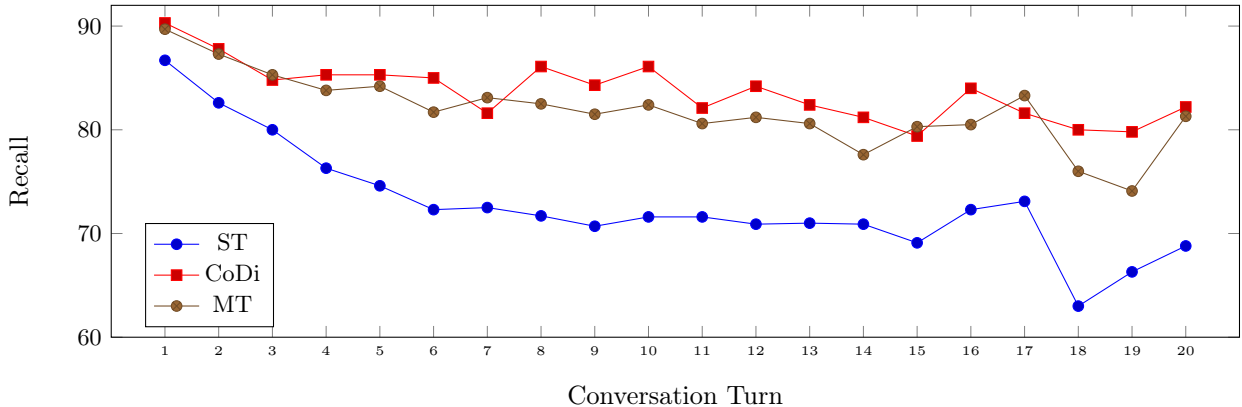
\begin{figure}[h]
        \centering
        \resizebox{\linewidth}{!}{
        \begin{tikzpicture}
            \begin{axis}[
                xlabel=Conversation Turn,
                ylabel=Recall,
                xmin=0, xmax=21,
                ymin=60, ymax=92,
                yticklabel style = {font=\small},
                xticklabel style = {font=\tiny},
                legend pos=south west,
                legend style = {font=\small},
                height=6cm,
                width=\linewidth,
                xtick={1,2,3,4,5,6,7,8,9,10,11,12,13,14,15,16,17,18,19,20},
                ]
            \addplot plot coordinates {
                (1, 86.7)
                (2, 82.6)
                (3, 80)
                (4, 76.3)
                (5, 74.6)
                (6, 72.3)
                (7, 72.5)
                (8, 71.7)
                (9, 70.7)
                (10, 71.6)
                (11, 71.6)
                (12, 70.9)
                (13, 71)
                (14, 70.9)
                (15, 69.1)
                (16, 72.3)
                (17, 73.1)
                (18, 63)
                (19, 66.3)
                (20, 68.8)
                };
                \addlegendentry{ST}
        
            \addplot plot coordinates {
                (1, 90.3)
                (2, 87.8)
                (3, 84.8)
                (4, 85.3)
                (5, 85.3)
                (6, 85)
                (7, 81.6)
                (8, 86.1)
                (9, 84.3)
                (10, 86.1)
                (11, 82.1)
                (12, 84.2)
                (13, 82.4)
                (14, 81.2)
                (15, 79.4)
                (16, 84)
                (17, 81.6)
                (18, 80)
                (19, 79.8)
                (20, 82.2)
                };
                \addlegendentry{CoDi}
                
            \addplot plot coordinates {
                (1, 89.7)
                (2, 87.3)
                (3, 85.3)
                (4, 83.8)
                (5, 84.2)
                (6, 81.7)
                (7, 83.1)
                (8, 82.5)
                (9, 81.5)
                (10, 82.4)
                (11, 80.6)
                (12, 81.2)
                (13, 80.6)
                (14, 77.6)
                (15, 80.3)
                (16, 80.5)
                (17, 83.3)
                (18, 76)
                (19, 74.1)
                (20, 81.3)
                };
                \addlegendentry{MT}
            \end{axis}
            \end{tikzpicture}}
    \caption{Average Per-Turn Model Recall on CoQA.}
    \label{fig:per_turn}
\end{figure}

In this exploration, we cover an important dimension to better understand our conversational distillation system for grounded reasoning: The per-turn performance on the CoQA evaluation dataset.
Specifically, we compare the per-turn test set performance of the gold single-turn, gold multi-turn and our full (1M samples) trained models. In Figure \ref{fig:per_turn} we find that our \methodname~Web model is on par with the gold multi-turn trained baseline for short conversations (three or less turns). For longer conversations, our approach near-consistently outperforms the multi-turn baseline. The single-turn baseline performs consistently worse, with the performance gap generally increasing in later turns.

\subsection{Language Modeling Evaluations}
We further evaluate our final model (based on the full 1M training process) on a commonly used subset of language model evaluations (LM evals) for small language models (e.g. used in \citet{liu2024mobilellm, allal2024SmolLM}). Namely, we evalute the performance for ARC-easy and -challenge \citep{clark2018think}, BoolQ \citep{clark-etal-2019-boolq}, PIQA \citep{bisk2020piqa}, SIQA \citep{sap2019social}, Hellaswag \citep{zellers2019hellaswag}, OBQA \citep{OpenBookQA2018} and Winogrande \citep{sakaguchi2021winogrande}. The goal of this evaluation is to show the performance of our grounded reasoning based distillation approach on this diverse set of language understanding tasks compared to pre-trained and instruction-tuned model alternatives. As shown in Table \ref{tab:lm_eval}, the \methodname~Web trained model on average still under performs instruction-tuned versions of the same base model, however, can improve over the pre-trained checkpoint, without any prompt adjustments during the evaluation. We believe that not regressing, yet even slightly improving the language model evaluation, shows promise of grounded reasoning tasks to act as rather generalist models.

\begin{table}[h]
    \centering
    \resizebox{\linewidth}{!}{
    \begin{tabular}{l|r|r|r|r|r|r|r|r|r}
        Model & Arc-E & Arc-C & BoolQ & PIQA & SIQA & Hellaswag & OBQA & Winogrande & Avg\\
        \hline
        Pre-Trained & 64.27 & 39.30 & 61.95 & \textbf{73.75} & 45.87 & 63.08 & 47.42 & 59.98 & 56.95 \\
        \methodname~Web & \textbf{65.38} & 38.83 & 67.97 & 73.54 & 46.19 & 61.60 & 47.66 & 60.47 & 57.71 \\
        Instruction-Tuned & 63.65 & \textbf{40.00} & \textbf{68.90} & 73.54 & \textbf{47.31} & \textbf{63.13} & \textbf{49.61} & \textbf{61.41} & \textbf{58.44} \\
    \end{tabular}
    }
    \caption{Language Model Evaluations of our CoDi trained student model compared to the same 1.4B Llama2-like base-model pre-trained and instruction-tuned.}
    \label{tab:lm_eval}
\end{table}

\subsection{Small-Scale Human Evaluation}
In this section, we present a small-scale human evaluation to confirm the validity of our evaluation metrics. We ask the human reviewers to rank three responses in the context of a textual document and the prior conversation (randomly taken from the CoQA test set). From the response ranking, we retrieve three binary win rates: (1) between our \methodname~Web predictions and gold labels, (2) comparing the human multi-turn trained baseline and gold labels, and (3) selecting our \methodname~Web predictions or the human multi-turn trained baseline. Table \ref{tab:human_eval_alignment} shows the results of the human evaluation\footnote{The double-blind human evaluation is executed by authors of the paper.} on 20 distinct documents containing 284 question answer rankings. We can see that there is a large overlap of equally good responses between our \methodname~distilled models and both, the multi-turn baseline and the gold answers. 
The results of the small-scale human evaluations 
further validate our findings using traditional metrics.

\begin{table}[h]
    \centering
    \begin{tabular}{l|l|r|r}
        Model 1 & Model 2 & Win Rate & Win+Tie Rate \\
        \hline
        \methodname~Web & Gold Annotations & 7.7 & 87.3 \\
        Human Multi Turn & Gold Annotations & 6.3 & 93.3 \\
        \methodname~Web & Human Multi Turn & 6.0 & 90.5 \\
    \end{tabular}
    \caption{Small-Scale Human Evaluation comparing \methodname against the human-annotated multi-turn trained model and gold labels.}
    \label{tab:human_eval_alignment}
\end{table}

\subsection{Zero-shot Summarization Performance}
Lastly, we explore another  zero-shot scenario in the realm of grounded reasoning tasks: abstractive summarization on CNN/DM \citep{nallapati2016abstractive} and XSum \citep{xsum-emnlp}. 
Table \ref{tab:summ_ablation} shows the zero-shot results obtained from instruction-tuned 1.4B and 7B Llama2-style baseline models, as well as the 3.8B Phi-3 checkpoint compared to our full \methodname~Web distilled model. We further show two supervised fine-tuned model comparisons to put the zero-shot performance into perspective, a fully fine-tuned 1.4B Llama2-style model and a competitive BART-Large model \citep{lewis2019bart}. Looking at the results, we find that \methodname~Web outperforms both Llama2-style instruction tuned baselines, at the 1.4B and 7B scale. While the Phi-3 checkpoint outperforms our model on the CNN-DM dataset, the delta is small given the significant size difference between the models. Comparing the models on the XSum dataset, \methodname~Web outperforms all other models, despite their significant size advantage.

\begin{table}[h]
    \centering
    \begin{tabular}{l|r|rrr|rrr}
    Eval Dataset & \#Params & \multicolumn{3}{c|}{CNN/DM} & \multicolumn{3}{c}{XSUM} \\
    \hline
    Metric & & R-1 & R-2 & R-L & R-1 & R-2 & R-L \\
    \hline
    Instruction-Tuned & 1.4B & 11.26 & 3.71 & 7.47 & 4.56 & 0.99 & 3.42 \\
    \methodname~Web & 1.4B & 24.71 & 7.73 & 17.36 & \textbf{17.66} & \textbf{2.60} & \textbf{13.49} \\
    Phi-3 & 3.8B & \textbf{28.48} & \textbf{9.04} & \textbf{18.05} & 12.59 & 2.59 & 9.04 \\
    Instruction-Tuned & 7B & 17.31 & 5.72 & 11.17 & 7.88 & 1.90 & 5.65 \\
    \hline
    Fine-Tuned & 1.4B & 41.03 & 17.42 & 29.59 & 28.14 & 10.95 & 23.04 \\
    BART large$\dagger$ & 406M & 44.16 & 21.28 & 40.90 & 45.14 & 22.27 & 37.25 \\  
\end{tabular}
    \caption{Abstractive summarization Zero-Shot Results on CNN/DM and XSum using our full \methodname~web model. Llama2 based, if no model mentioned. $\dagger$ = Results taken from \cite{lewis2019bart}}
    \label{tab:summ_ablation}
\end{table}

\section{Related Work}

\paragraph{High Quality Data Distillation}
Recently, there has been a large push towards curating high quality datasets for training small language models. With the intuition that small models are more sensitive to low-quality data, recent research (1) filtered datasets based on quality, (2) rewrites data samples to improve quality, and (3) synthesizes new and diverse samples to teach the model the desired behavior. 
For example, in their seminal work, \citet{gunasekar2023textbooks} use code data from the web and refine it to ``textbook-style'' samples to pre-train small scale-decoder only models. 
Similarly, \citet{zhou2023lima} argue for the need of high-quality datasets, even for alignment purposes. Along similar lines, \citet{wei2022finetuned, longpre2023flan} show that dataset diversity along the task axis plays a crucial role for model training.
In the creative writing domain, \citet{ravi2024small} show that small language models are able to learn difficult concepts, such as humor, when distilled in an interactive manner.

\paragraph{Conversational Question Answering}
Conversational question answering has been explored extensively in a variety of works given the importance of the task. For example, \citet{liu2024chatqa} propose a family of conversational question answering models at large scales by adding a dense retrieval module. 
\citet{feng-etal-2020-doc2dial} propose a new method to create conversational datasets using discourse units, while
\citet{anantha-etal-2021-open} proposes a question rewriting method in the conversational context.
\citet{adlakha-etal-2022-topiocqa} publish a new dataset for conversational question answering focusing on topic switches. Compared to these approaches, our new conversational synthesis approach is more scalable, while maintaining data diversity and steerability.

\paragraph{Prompting Paradigms}
In black-box LLM distillation, the human curated prompt plays a major role for the downstream performance. While many different approaches have been proposed in the past, so called 
``Chain-of-Thought'' (CoT) prompting is one of the most popular black-box LLM distillation paradigms to achieve high-quality results \citep{wei2023chainofthought}. To this end, we follow the approach taken in in \citet{wei2023chainofthought} and prompt our per-turn conversational links using a flavor for CoT prompting, asking the model to produce a reasoning trace in CoT style.

\paragraph{Small Language Models}
Given the strong generalist performance of LLMs, such as the GPT \citep{openai2024gpt4} and the Llama \citep{touvron2023llama} series, the question on how much these abilities can be distilled into SLMs has become important research question.
For example, in the Orca work, \citet{mukherjee2023orca} show  promising performance at the 13B model scale when distilling data from GPT-4 using explanation traces. Similarly, the Phi series \citep{gunasekar2023textbooks} shows strong performance of even smaller language models when trained on code data. Lastly, OpenELM \citep{mehta2024openelm} shows similar results.

\section{Conclusion}
In this paper we show a novel conversational distillation method applied to the challenging task of conversational grounded reasoning for question answering. We show that using our framework to generate diverse, steerable and conversational question answer traces can significantly close the intra-domain performance gap compared to human curated multi-turn conversations. Furthermore, we show that our synthesis approach can improve zero-shot question answering and summarization performance compared to similar sized instruction-tuned models, and even outperform models of significantly larger size. With these promising results, we make a compelling case for using the \methodname~framework to synthesize data from diverse seeds instead of going through the resource-intensive human annotation process or scaling up the model size.

\newpage
\section{Contributions}
\begin{table}[h]
    \centering
    \begin{tabular}{c|c|c|c|c|c|c|c|c|c}
        Contributor & Patrick & Arash & Rylan & Kanika & Matt & Waqar & Adithya & Ahmed & Akshat \\
        \hline
        Conceptualization &$\bullet$&&&&&&&&$\bullet$ \\
        \hline
        Experimentation &$\bullet$&$\bullet$&&$\bullet$&$\bullet$&&&&$\bullet$ \\
        \hline
        Applications & $\bullet$ & $\bullet$ & $\bullet$ & $\bullet$ & $\bullet$ &$\bullet$&&& $\bullet$ \\
        \hline
        Literature Review & $\bullet$ &$\bullet$&&&&&&&$\bullet$ \\
        \hline
        Paper Writing & $\bullet$ &$\bullet$&$\bullet$&$\bullet$&&&&&$\bullet$ \\
        \hline
        Leadership &&&&&&&$\bullet$&$\bullet$&$\bullet$ \\
    \end{tabular}
    \caption{Author Contribution Matrix}
    \label{tab:my_label}
\end{table}


\clearpage
\newpage
\bibliographystyle{assets/plainnat}
\bibliography{paper,anthology,custom}


\end{document}